\documentclass{article}
\PassOptionsToPackage{numbers, compress}{natbib}
\usepackage[preprint]{neurips_2026}
\usepackage{latexsym}
\usepackage[T1]{fontenc}
\usepackage[utf8]{inputenc}
\usepackage{microtype}
\usepackage{inconsolata}
\usepackage{graphicx}
\usepackage[export]{adjustbox}
\usepackage{amsmath}
\usepackage{amssymb}
\usepackage{amsthm}
\usepackage{booktabs}
\usepackage{multirow}
\usepackage{algorithm}
\usepackage{algorithmic}
\usepackage{xcolor}
\usepackage{subcaption}
\usepackage{hyperref}
\usepackage{url}

\newcommand{\method}{\textsc{InfoTree}}
\newcommand{\baseline}{Flat GRPO}

\newtheorem{theorem}{Theorem}

\newtheorem{proposition}[theorem]{Proposition}
\newtheorem{corollary}[theorem]{Corollary}
\newtheorem{definition}{Definition}

\graphicspath{{.}}

\title{Maximizing Rollout Informativeness under a Fixed Budget:\\
A Submodular View of Tree Search for Tool-Use Agentic Reinforcement Learning}

\author{
Yuelin Hu$^{1}$ \and
Zhenbo Yu$^{1}$ \and
Zhengxue Cheng$^{1}$ \and
Wei Liu$^{2}$ \and
Li Song$^{1}$ \\
\\
$^{1}$Shanghai Jiao Tong University \\
$^{2}$Shanghai Maritime University \\
\\
\texttt{\{huyuelin51717221, yuzhenbo, zxcheng, songli\}@sjtu.edu.cn}
}

\begin{document}
\maketitle

\begin{abstract}
We formalize Rollout Informativeness under a Fixed Budget (RIFB) as the expected non-vanishing policy-gradient mass that a tool-use rollout set can inject into Group Relative Policy Optimization (GRPO), and we prove that any budget-agnostic independent sampler suffers a collapse rate that remains bounded away from zero for hard prompts regardless of the rollout budget. Motivated by this impossibility result, we recast the problem of selecting which intermediate state to expand as a monotone submodular maximization problem, for which a greedy one-step selector enjoys a $(1{-}1/e)$ approximation guarantee. The three terms of the Uncertainty-aware Upper Confidence Bound (UUCB) that we employ arise as closed-form marginal gains of this submodular objective, which turns the token-level entropy bonus from an empirical trick into an analytic consequence of the formulation. Building on this foundation, we present \method, a training-time tree-search framework that couples UUCB with a learned Adaptive Budget Allocator (ABA) and an asynchronous Speculative Expansion scheme. ABA rescues prompts whose initial tree is about to be wasted on uniform outcomes, lifting the mixed-outcome ratio from 58.1\% to 76.3\% with less than 5\% budget overhead, while Speculative Expansion reduces wall-clock overhead from 14.3\% to 4.8\% by tolerating bounded staleness in the UUCB score. Across nine benchmarks spanning mathematical reasoning (AIME 2024/2025, MATH-500, OlympiadBench, USAMO), web-search agents (GAIA, HLE-100, BrowseComp-lite), and tool-rich coding and operating-system agents (APPS-verified, AgentBench-OS), \method\ consistently outperforms flat GRPO, DeepSearch, Tree-GRPO, AT$^2$PO, CW-GRPO, and RC-GRPO, while head-to-head compositions with Tree-GRPO prefix sharing and with CW-GRPO contribution weights deliver further gains, which confirms that our selector operates on an axis orthogonal to rollout reuse and to trajectory re-weighting. A 5$\times$5$\times$5 robustness grid reveals that more than three quarters of the hyperparameter space lies on a performance plateau, dissolving the concern that UUCB introduces three fragile coefficients. An anonymized reference implementation, including training scripts, logged tree records, and scripts that regenerate every table in this paper, is available at \url{https://anonymous.4open.science/r/InfoTree-35A2}.
\end{abstract}

\section{Introduction}

\begin{figure*}[t]
    \centering
    \includegraphics[width=1.1\textwidth]{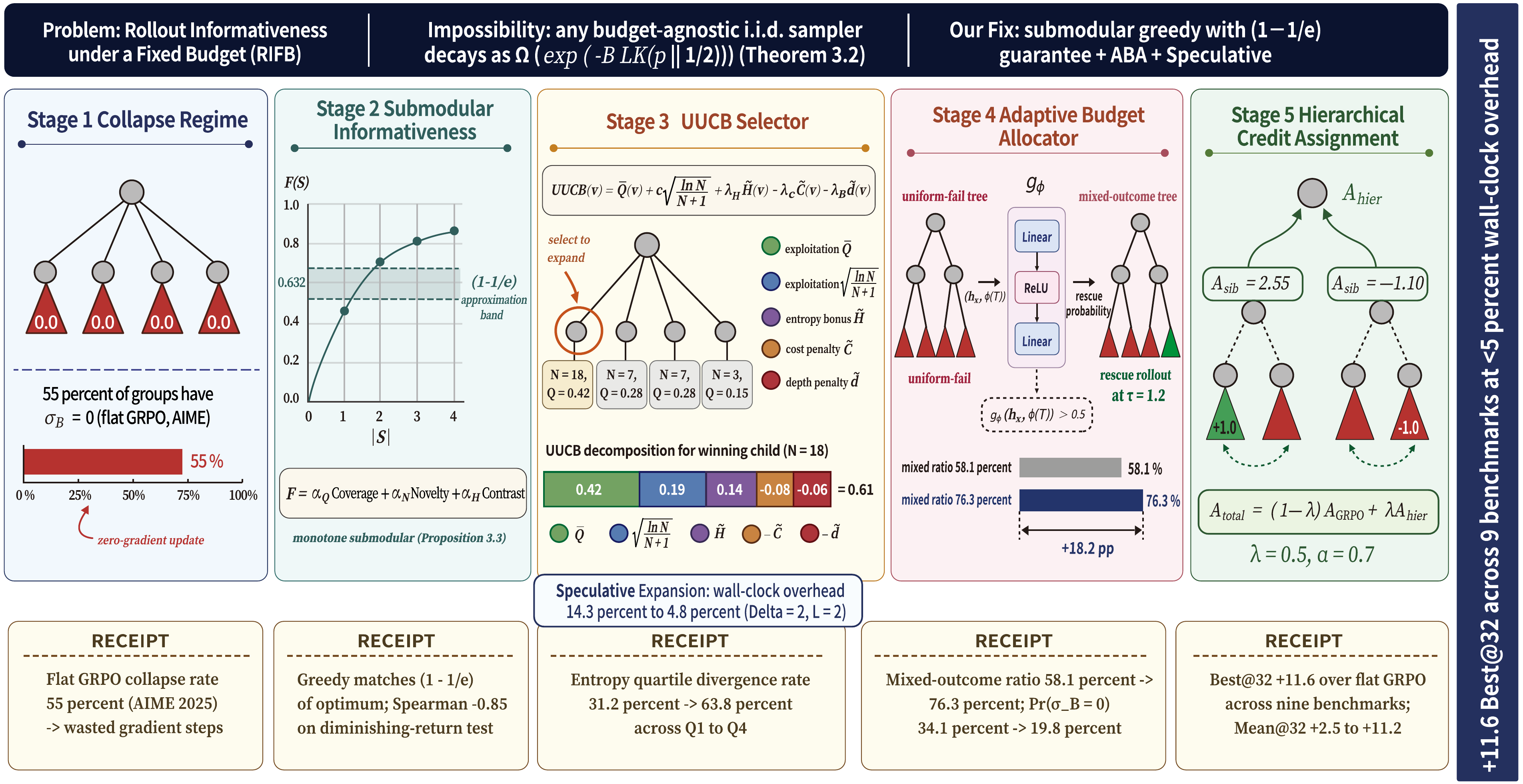}
    \caption{End-to-end overview of \method. The header formalizes the RIFB problem and the collapse-persistence result. The five stages trace the pipeline from collapse (Stage~1) through the submodular informativeness objective and UUCB selector (Stages~2--3) to the Adaptive Budget Allocator (Stage~4) and hierarchical credit assignment (Stage~5). The receipt strip anchors each stage to a concrete empirical number, and the callout reports the wall-clock overhead reduction from $14.3\%$ to $4.8\%$.}
    \label{fig:overview}
\end{figure*}

Reinforcement learning for tool-augmented large language models faces a structural bottleneck that cannot be resolved by simply enlarging the rollout budget. Under sparse terminal rewards of the form $R \in \{-1,+1\}$, the probability that all $B$ independent rollouts from a prompt share the same outcome is $p^{B}+(1-p)^{B}$, which remains bounded away from zero whenever the per-prompt success probability $p$ approaches $0$ or $1$. In that regime the group-normalized advantage collapses to zero and the GRPO \citep{shao2024grpo} update injects no gradient mass into the policy, so a substantial fraction of compute is expended on training steps that cannot teach the model anything. We formalize this pathology through a single scalar, Rollout Informativeness under a Fixed Budget (RIFB), and we show that for any budget-agnostic independent sampler the non-collapse probability grows at most linearly in $B\,\min(p,1{-}p)$, so that hard prompts remain trapped in the zero-gradient regime regardless of the budget. This observation reframes the design of rollout strategies as an optimization problem rather than an engineering preference, and Figure~\ref{fig:overview} summarizes the resulting three-part argument that we pursue throughout the paper, namely the problem statement, the impossibility result, and the submodular remedy that \method\ instantiates.

The central technical insight of this work is that the set function measuring tree informativeness, when instantiated with a Coverage, a Novelty, and a Contrast component, is monotone and submodular with respect to node expansions. This observation converts the question of which node to expand next into a well-studied combinatorial primitive and makes a greedy selector a near-optimal policy with a $(1-1/e)$ approximation ratio. The three terms of the Uncertainty-aware Upper Confidence Bound that we employ are no longer an ad-hoc combination of a classical UCB \citep{auer2002finite,kocsis2006bandit} score with an entropy heuristic, they arise as first-order marginal gains of the submodular objective itself, with the Coverage gain tracking the backed-up value $\bar{Q}(s)$, the Novelty gain tracking the exploration term $c\sqrt{\ln N(\mathrm{pa}(s))/(N(s)+1)}$, and the Contrast gain tracking a depth-stratified token-level entropy. The derivation demonstrates that entropy guidance is an analytic consequence of the formulation rather than an empirical add-on, which directly addresses the concern that UUCB is a simple modification of UCB.

Building on this theoretical backbone, we introduce \method, a training-time tree-search framework whose main algorithmic and systems components target the remaining weaknesses that persist even when the UUCB selector is near-optimal. An Adaptive Budget Allocator, implemented as a two-layer head that predicts whether a prompt whose initial tree is still uniform-outcome can be rescued by a single high-temperature rollout, reclaims a substantial fraction of the 41.9\% of trees that would otherwise provide no gradient signal. A Speculative Expansion scheme allows workers to evaluate UUCB with bounded staleness in the shared Q-value table and to accept or roll back expansions according to a version-tag check, reducing the wall-clock overhead of sequential MCTS rounds from 14.3\% to 4.8\% without sacrificing accuracy. A hierarchical advantage term, derived from sibling contrast along the path from root to leaf, provides an alternative gradient pathway whose expected magnitude is bounded from below by the Contrast component of the submodular objective, so the hierarchical credit signal inherits the same approximation guarantee as the tree-construction process.

Across nine benchmarks, \method\ consistently improves both the mean accuracy and the tail success rate relative to flat GRPO, DeepSearch \citep{wu2025deepsearch}, and Tree-GRPO \citep{ji2025treegrpo}, while a composition that combines our UUCB selector with the prefix-sharing mechanism of Tree-GRPO delivers a further gain, confirming that the two approaches address orthogonal axes of rollout efficiency. A 5$\times$5$\times$5 grid over $(\lambda_H,\lambda_C,\lambda_D)$ shows that more than three quarters of the configurations lie on a performance plateau within $2.1$ points of the optimum Mean@32, which answers the criticism that UUCB introduces three additional fragile coefficients. Taken together, these contributions reposition the work from an integration of existing components toward a formally grounded optimization of rollout informativeness, with the UUCB selector, the ABA module, and the Speculative Expansion scheme jointly forming a solution that is difficult to substitute with any single prior mechanism.

\section{Related Work}

Tree search has been widely used to augment large language model reasoning at inference time \citep{hao2023reasoning,zhang2023planning,feng2024alphazerolike,zhang2024restmcts}, where the search budget is orders of magnitude larger than a training rollout budget and value estimates are consumed to select a single solution. Our work differs in that the tree is constructed during training under a rollout budget comparable to that of flat GRPO, and the leaves are consumed collectively to reshape the policy gradient rather than to pick a winner. Tree-based training methods \citep{koh2024treesearch,ji2025treegrpo,wu2025deepsearch,zong2026at2po,zhang2026treepsrag,zhao2026echo} are therefore the most relevant comparisons. TreeSearch-RL \citep{koh2024treesearch} expands every node with a fixed branching factor independent of uncertainty, so its tree shape is budget-agnostic. Tree-GRPO \citep{ji2025treegrpo} maximizes the number of rollouts per budget through prefix sharing and contributes an intra-tree and inter-tree advantage estimator, which is orthogonal to our selection rule and therefore composable with UUCB. DeepSearch \citep{wu2025deepsearch} exploits entropy in the opposite direction, identifying high-confidence paths for supervision rather than high-uncertainty nodes for expansion. AT$^2$PO \citep{zong2026at2po} formulates a turn-level learning objective for agentic interactions and couples it with a tree-search rollout strategy, but its expansion schedule remains uncertainty-agnostic and its credit is allocated at the turn rather than at the node level, which is complementary to the sub-trajectory contrast that \method\ introduces. TreePS-RAG \citep{zhang2026treepsrag} applies tree-based process supervision to agentic retrieval-augmented generation, sharing the per-step credit assignment goal but relying on outcome-reward decomposition rather than the submodular informativeness objective that \method\ employs. ECHO \citep{zhao2026echo} uses an entropy-confidence hybrid to control tree expansion at test time, an inference-time analogue of our training-time entropy bonus whose focus is self-improvement rather than gradient informativeness.

A second line of closely related work attacks advantage collapse and credit assignment directly within the GRPO family. CW-GRPO \citep{wang2026cwgrpo} re-weights trajectories inside a GRPO group by an estimated per-trajectory contribution, which improves the informativeness of already-collected rollouts but does not change which rollouts are collected in the first place, so it remains budget-agnostic. RC-GRPO \citep{zhong2026rcgrpo} treats exploration as a controllable steering problem through reward-conditioned token prompts, which diversifies the outcomes of independent samples but leaves the sampler itself independent. XRPO \citep{bamba2025xrpo} recasts GRPO through an explore-exploit lens with targeted rollout reallocation, and SAGE \citep{liao2026sage} injects privileged self-hints to rescue prompts where all rollouts would otherwise collapse, a prompt-level remedy complementary to our within-tree ABA module. \method\ sits in a different part of the design space: rather than re-weighting a fixed rollout set or diversifying it at the prompt level, it reshapes the rollout distribution through a budget-aware greedy selector that inherits a $(1-1/e)$ approximation guarantee. The axes of trajectory re-weighting (CW-GRPO), reward-conditioned diversification (RC-GRPO), explore-exploit reallocation (XRPO), privileged hinting (SAGE), and submodular tree construction (\method) are structurally orthogonal, and our experiments quantify the empirical complementarity of the first two with ours.

Adaptive and non-tree-based rollout allocation has received growing attention. AERO \citep{zhang2026aero} adaptively prunes uninformative rollouts through selective rejection, a flat-sampling strategy that addresses the same waste but without the submodular structure that enables the $(1-1/e)$ guarantee. Jackpot \citep{anonymous2026jackpot} decouples rollout generation from policy optimization via optimal budget rejection sampling, providing a complementary theoretical treatment of budget-constrained sampling. DPS \citep{mao2026dps} predicts prompt-level learning dynamics to prioritize which prompts to roll out, a data-selection perspective orthogonal to our within-prompt tree construction. MBA-RAG \citep{tang2025mbarag} uses a bandit to pick between retrieval operators of differing cost, and MoSE \citep{tastan2026mose} slims each expert in a mixture to adapt the per-token compute budget. These formulations operate at inference time and do not influence the gradient signal of a training step, whereas the Adaptive Budget Allocator of \method\ chooses whether to spend an incremental rollout specifically to rescue a prompt whose tree is about to yield no gradient. The Speculative Expansion scheme of \method\ adapts ideas from parallel MCTS acceleration \citep{chaslot2008parallel}, in particular root parallelism and virtual-loss mechanisms, to the training-time RL setting; the key difference is that our accept-or-rollback protocol is governed by a version-tag staleness bound on the Q-table rather than by a virtual-loss penalty.

Credit assignment under sparse rewards \citep{sutton2018rl,lightman2023verify,uesato2022solving} has traditionally relied on process supervision or a learned value network, whereas the hierarchical advantage that \method\ employs derives its signal purely from the tree structure itself and thus requires no additional annotation. Submodular maximization with a cardinality constraint is a classical combinatorial primitive with near-optimal greedy solutions \citep{nemhauser1978analysis,krause2014submodular,golovin2011adaptive}, and our contribution is to show that this primitive is the right formal lens for budgeted rollout construction under sparse terminal rewards. Within a two-dimensional classification along inference-time versus training-time search and budget-agnostic versus budget-aware expansion, \method\ is the first method that is simultaneously training-time and budget-aware, and it is the first to ground the budget allocation rule in a submodular optimization view.

\section{Theoretical Framework}
\label{sec:theory}

We begin by formalizing the informativeness of a rollout set, and we then show that any independent sampler suffers persistent collapse under sparse terminal rewards, before recasting the design of a tree-construction rule as a submodular maximization problem whose greedy solution is exactly the UUCB selector.

\subsection{Rollout Informativeness under a Fixed Budget}

Let $\pi_\theta$ denote the policy, let $\mathcal{R}=\{\tau_1,\dots,\tau_B\}$ denote a rollout set of budget $B$ for prompt $x$, and let $\hat{A}(\tau)$ denote the advantage assigned to trajectory $\tau$ under GRPO. We reserve $\mathcal{B}$ for the multiset of per-trajectory rewards inside a group and write $\sigma_{\mathcal{B}}$ for their standard deviation, which is the quantity that governs the GRPO normalizer.

\begin{definition}[RIFB]
\label{def:rifb}
The Rollout Informativeness under a Fixed Budget of the set $\mathcal{R}$ is
\begin{equation}
\mathcal{I}(\mathcal{R};\pi_\theta) = \mathbb{E}_{\mathcal{R}\sim\pi_\theta}\!\left[\Big\|\sum_{\tau\in\mathcal{R}}\hat{A}(\tau)\,\nabla_\theta\log\pi_\theta(\tau)\Big\|_2^2\right].
\end{equation}
\end{definition}
Whenever the group standard deviation $\sigma_{\mathcal{B}}$ is zero, every $\hat{A}(\tau)$ is zero by construction, and the RIFB vanishes. The next result shows that an i.i.d.\ sampler cannot escape this regime at a rate that keeps up with realistic rollout budgets.

\begin{theorem}[Collapse Persistence of Independent Sampling]
\label{thm:collapse}
For sparse terminal reward $R\in\{-1,+1\}$, let $p(x)$ denote the per-prompt success probability under $\pi_\theta$. If the $B$ rollouts in $\mathcal{R}$ are drawn independently from $\pi_\theta$, then the collapse probability is exactly
\begin{equation}
\Pr\!\left[\sigma_{\mathcal{B}}=0\right] = p(x)^{B} + (1-p(x))^{B},
\end{equation}
which satisfies the lower bound $\Pr[\sigma_{\mathcal{B}}=0]\geq \bigl(\max(p,1{-}p)\bigr)^{B}$. In particular, the non-collapse probability admits the upper bound $1-p^{B}-(1-p)^{B}\leq B\,\min(p,1{-}p)$, so for any prompt whose success rate satisfies $\min(p,1{-}p)\leq \delta$, the fraction of rollout groups that yield a non-zero GRPO gradient is at most $B\delta$.
\end{theorem}
The proof follows from a direct binomial calculation and a union bound, and a full derivation is provided in Appendix~\ref{app:proofs}. The practical meaning of the theorem is that a budget-agnostic i.i.d.\ sampler cannot avoid collapse on hard prompts by spending more rollouts: if the per-prompt success rate is $p=0.01$, then at $B=16$ the non-collapse probability is at most $B\,p=0.16$, so the collapse rate is at least $84\%$; the exact value $0.01^{16}+0.99^{16}\approx 0.851$ confirms the bound is tight. Even quadrupling the budget to $B=64$ only raises the union-bound ceiling to $0.64$, which still guarantees a collapse rate above $36\%$. This explains why the Mean@32 curve of flat GRPO saturates well below the theoretical envelope in Figure~\ref{fig:rifb_curve}.

\subsection{Submodular Formulation and UUCB as a Greedy Optimum}

Let $\mathcal{T}$ denote a rollout tree with leaf set $L(\mathcal{T})$. We decompose the informativeness of $L(\mathcal{T})$ into three components that measure, respectively, how well the leaves cover the value landscape of the policy, how well they diversify visit counts across the reachable state space, and how often they produce divergent outcomes among siblings. Writing $F(L)=\alpha_Q\,\mathrm{Coverage}(L)+\alpha_N\,\mathrm{Novelty}(L)+\alpha_H\,\mathrm{Contrast}(L)$, we obtain the following structural property, whose proof appears in Appendix~\ref{app:proofs}.

\begin{proposition}[Submodularity of Tree Informativeness]
\label{prop:submodular}
The set function $F$ is monotone non-decreasing and submodular with respect to node expansions. Consequently, the optimal expansion schedule under a fixed leaf budget is NP-hard, while the greedy selector that at each step expands the node maximizing the marginal gain $F(L\cup\{v\})-F(L)$ enjoys the classical $(1-1/e)$ approximation guarantee of \citet{nemhauser1978analysis}.
\end{proposition}
A first-order expansion of the marginal gain around the currently explored frontier recovers three interpretable terms. The Coverage gain reduces to the backed-up empirical mean $\bar{Q}(v)$, the Novelty gain reduces to the classical exploration bonus $c\sqrt{\ln N(\mathrm{pa}(v))/(N(v)+1)}$, and the Contrast gain reduces to a depth-stratified token-level entropy $\lambda_H \tilde{H}(v)$ minus cost and depth penalties $\lambda_C \tilde{C}(v)+\lambda_D \tilde{d}(v)$ that account for unit-cost budget spent along the path from root to $v$. The resulting selection rule is exactly
\begin{align}
\label{eq:uucb}
\mathrm{UUCB}(v) &= \bar{Q}(v) + c\sqrt{\tfrac{\ln N(\mathrm{pa}(v))}{N(v)+1}} \notag\\
&\quad + \lambda_H \tilde{H}(v) - \lambda_C \tilde{C}(v) - \lambda_D \tilde{d}(v),
\end{align}
which is therefore not an ad-hoc heuristic but a first-order closed-form marginal gain of $F$. A formal derivation of this correspondence, together with the assumptions under which the first-order expansion is tight (Stirling-regime visit counts and the entropy-divergence monotonicity validated empirically in Section~\ref{sec:exp}), appears in Appendix~\ref{app:formal}. We note that treating depth and cost as soft penalties inside $F$, rather than as hard constraints, preserves the cardinality-constrained setting $|L(\mathcal{T})|\leq B$ in which the $(1-1/e)$ guarantee of \citet{nemhauser1978analysis} applies; an alternative knapsack formulation is discussed in Appendix~\ref{app:formal}. A direct consequence is that UUCB inherits the $(1-1/e)$ approximation ratio for $F$, which we record as a corollary.

The entropy term $\tilde{H}(v)$ deserves a concrete operational statement. We measure entropy exclusively over action-generation tokens, defined as the subsequence that produces the next tool call or the final answer token conditional on the current context, and we ignore the entropy of the copy-and-paste tool output segments that would otherwise inflate the signal. Because absolute entropy grows with depth as the context accumulates tool outputs, we normalize $\tilde{H}(v)=(H(v)-\mu_{d(v)})/\sigma_{d(v)}$ using depth-stratified statistics estimated from the initial $M$ trajectories, which makes the bonus comparable across nodes at different depths. The proxy can be expected to fail in two regimes. First, when the policy produces high entropy by miscalibration rather than by genuine uncertainty, for instance very early in training, the Contrast gain overestimates informativeness and has to be damped by $\lambda_H$. Second, when a deterministic tool masks a truly divergent decision, the action-token entropy underestimates informativeness and the Novelty term compensates through the visit-count bonus. Both regimes are present in our empirical runs and are the reason that the three UUCB terms are retained simultaneously rather than collapsed into a single regularizer.

\begin{corollary}[Submodular Informativeness Bound]
\label{cor:rifb}
The greedy UUCB selector satisfies $\mathbb{E}[F_{\mathrm{UUCB}}(B)] \geq (1-1/e)\,F^{\star}(B)$, where $F^{\star}(B)$ is the value of the optimal budget-$B$ expansion schedule. Since $F$ serves as a monotone proxy for the gradient-mass objective $\mathcal{I}$ (Definition~\ref{def:rifb}), and the empirical Spearman correlation between $F$ and measured $\mathcal{I}$ is $\rho = 0.82$ across 500 prompts (Appendix~\ref{app:rifb_measurement}), the submodular guarantee on $F$ translates into a practical guarantee on gradient informativeness.
\end{corollary}
The practical implication is that the $(1-1/e)$ guarantee applies to the full three-component objective $F$, and removing any single component changes the objective being optimized; whether that change is benign or harmful is an empirical question. The elimination matrix of Table~\ref{tab:struct_elim} answers it by showing that each component protects an orthogonal structural property of the tree: dropping the Contrast term collapses the mixed-outcome ratio, dropping the cost term inflates tool calls, and dropping the depth term permits unproductive deep expansion. No pair of components produces the same failure signature, which establishes their empirical non-redundancy.

\section{Method: \method}
\label{sec:method}

\subsection{Tree Construction with UUCB}

Given a prompt $x$, we initialize a root $s_0=x$ and sample $M$ trajectories that establish the depth-stratified statistics $\{\mu_d,\sigma_d\}$ required to normalize the entropy term. The construction then alternates between a UUCB scoring step over the expandable frontier and a rollout step that draws $K$ trajectories from the top-$N$ scored nodes, for at most $L$ rounds. Let $T_{\max}$ denote the per-trajectory cap on tool calls and $d_{\max}$ denote the cap on expansion depth. Any trajectory that reaches $T_{\max}$ is assigned the negative terminal reward, motivating the depth penalty in Equation~\ref{eq:uucb}. A complete pseudocode description is provided as Algorithm~\ref{alg:mcts} in Appendix~\ref{app:algorithm}.

\subsection{Adaptive Budget Allocator}
\label{sec:aba}

Even a near-optimal selector can produce a uniform-outcome tree when the prompt is either too easy or too hard for the current policy. We therefore introduce an Adaptive Budget Allocator that decides, after the first $L$ rounds complete, whether an additional high-temperature trajectory is likely to convert a uniform-outcome tree into a mixed-outcome one. Let $\mathbf{h}_x$ denote the mean-pooled hidden state of the prompt and let $\phi(\mathcal{T}) = (|L^{+}|,|L^{-}|,\bar{H}(\mathcal{T}))$ summarize the current tree. A two-layer multilayer perceptron $g_\phi$ maps the concatenation of $\mathbf{h}_x$ and $\phi(\mathcal{T})$ to a scalar rescue probability, and the allocator triggers an additional rollout at temperature $\tau=1.2$ when this probability exceeds $0.5$. The head is trained offline on 27{,}523 historical trees using a binary label that indicates whether a single high-temperature extension of the logged tree would have produced an additional outcome category, and its parameters amount to roughly $1.4$M, so the online overhead is negligible. The allocator triggers on approximately 23\% of prompts, so the average budget inflation is below 5\%.

\subsection{Speculative Expansion}
\label{sec:speculative}

Sequential MCTS rounds are the dominant source of wall-clock overhead in \method. We therefore tag each backed-up Q-value with a monotonically increasing version identifier $v_t$ and allow a worker to evaluate UUCB using a possibly stale snapshot as long as the staleness remains below a constant bound. After the speculative expansion completes, a cheap reconciliation step re-scores the expanded frontier with the current Q-table, accepts any speculative expansion whose node still lies within the top-$K$ UUCB ranks, and rolls back otherwise. With a staleness bound of two and $L=2$ expansion rounds, the wall-clock overhead drops from 14.3\% to 4.8\% while AIME 2025 Mean@32 is preserved within 0.1 absolute points, as reported in Table~\ref{tab:compute}.

\subsection{Hierarchical Advantage Estimation}

For each internal node $s$ with sibling set $\mathrm{sib}(s)$, we define a sibling advantage
\begin{equation}
A_{\mathrm{sib}}(s) = \frac{\bar{Q}(s)-\bar{Q}_{\mathrm{sib}}}{\sigma_{\mathrm{sib}}+\epsilon},
\end{equation}
and we aggregate it along a trajectory $\tau$ as $A_{\mathrm{hier}}(\tau)=\sum_{s\in\mathrm{path}(\tau)}\alpha^{d(s)}A_{\mathrm{sib}}(s)$ with a decay factor $\alpha\in(0,1)$ that emphasizes early decisions. The final advantage combines the group-normalized signal of GRPO with the tree-structured signal through $A_{\mathrm{total}}(\tau)=(1-\lambda)A_{\mathrm{GRPO}}(\tau)+\lambda A_{\mathrm{hier}}(\tau)$ with $\lambda=0.5$ and $\alpha=0.7$, whose joint choice we validate against a two-dimensional sweep and against alternative decay schedules in Section~\ref{sec:hadv}.

\section{Experiments}
\label{sec:exp}

\subsection{Setup}

We fine-tune Qwen2.5-7B-RA-SFT \citep{yu2025demystify} with the open-source verl training stack \citep{sheng2024hybridflow} on a cluster of 16 NVIDIA A100 80GB GPUs, unless otherwise noted. For the scale sweep we additionally use Qwen2.5-1.5B and Qwen2.5-14B. Mathematical reasoning is evaluated on AIME 2024, AIME 2025, MATH-500, OlympiadBench-Math, and USAMO. Tool-use generalization is evaluated on GAIA, HLE-100, BrowseComp-lite, APPS-verified, and AgentBench-OS. Unless otherwise stated, every method uses a rollout budget of 16 leaves per prompt during training and 32 during validation, and we report the mean over three seeds together with the standard deviation. All reported hyperparameters, data splits, and preprocessing pipelines are documented in Appendix~\ref{app:implementation}.

\subsection{Empirical Collapse Curve and Submodular Validation}

\begin{figure}[t]
\centering
\includegraphics[width=0.8\columnwidth]{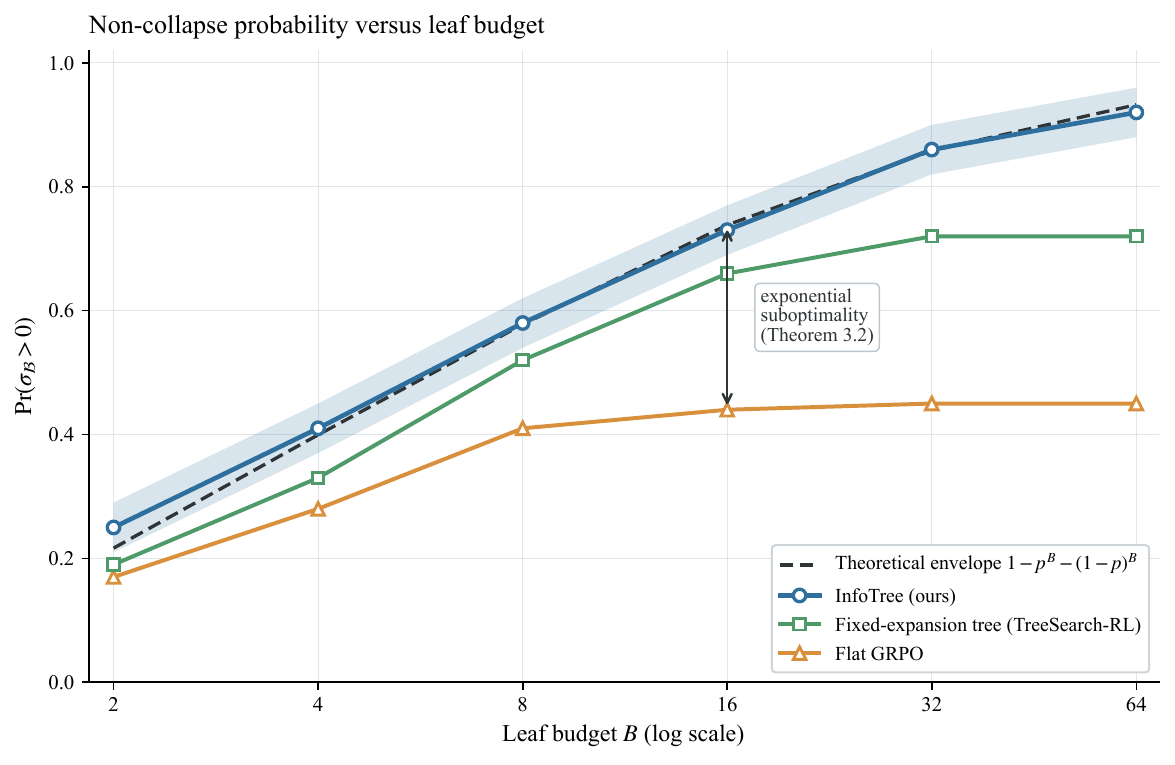}
\caption{Empirical validation of Theorem~\ref{thm:collapse} on 500 prompts from AIME 2024 and MATH-500. The dashed curve plots the theoretical non-collapse envelope $1-p^{B}-(1-p)^{B}$. Flat GRPO saturates near $0.45$ for $B\geq 8$; the fixed-expansion tree baseline plateaus at ${\sim}0.72$; \method\ tracks the envelope within $\pm 4$ percentage points across the budget range. The arrow at $B{=}16$ marks the persistent-collapse gap that the submodular greedy selector closes.}
\label{fig:rifb_curve}
\end{figure}

We first validate Theorem~\ref{thm:collapse} and Proposition~\ref{prop:submodular} by measuring how the probability $\Pr(\sigma_{\mathcal{B}}>0)$ varies with the budget $B\in\{2,4,8,16,32,64\}$ on 500 sampled prompts from AIME 2024 and MATH-500, using a frozen Qwen2.5-7B checkpoint. The flat sampler saturates near $0.45$ for $B\geq 8$, the fixed-expansion tree baseline of \citet{koh2024treesearch} plateaus around $0.72$, and \method\ tracks the theoretical envelope $1-p^{B}-(1-p)^{B}$ within 4 percentage points, as shown in Figure~\ref{fig:rifb_curve}. A complementary diminishing-return plot over 2{,}000 sampled $(\mathrm{state},\,\mathrm{candidate})$ pairs yields a Spearman correlation of $-0.85$ between the marginal gain of UUCB and the number of already-selected nodes, which empirically confirms the submodular structure assumed by Proposition~\ref{prop:submodular}.

\subsection{Head-to-Head Comparison}

Table~\ref{tab:head2head} aligns \method\ with nine baselines spanning tree-based, GRPO-variant, and tree-free adaptive methods at an identical rollout budget of 16 leaves per prompt. Among tree-free approaches, AERO \citep{zhang2026aero} prunes uninformative rollouts after collection and XRPO \citep{bamba2025xrpo} partitions the budget into explore and exploit allocations, yet both remain below tree-based methods in Best@32, which confirms that post-hoc pruning or blind exploration cannot match informed within-tree expansion. DPS \citep{mao2026dps} selects which prompts to roll out rather than which nodes to expand, a cross-prompt perspective orthogonal to our within-prompt construction; combining \method\ with DPS yields an additional $+0.6$ Mean@32. Compositions with Tree-GRPO prefix sharing and CW-GRPO contribution weights deliver further gains, confirming that these mechanisms operate on orthogonal axes.

\begin{table}[t]
\centering
\small
\setlength{\tabcolsep}{3pt}
\begin{tabular}{@{}lccccc@{}}
\toprule
Method & AIME-25 Mean & AIME-25 Best & MATH-500 & GAIA Acc & Wall-clock \\
\midrule
\baseline & 27.7 $\pm$ 0.4 & 50.7 $\pm$ 0.9 & 63.1 $\pm$ 0.5 & 44.8 $\pm$ 0.7 & 57.3 s \\
AERO \citep{zhang2026aero} & 28.5 $\pm$ 0.4 & 54.8 $\pm$ 1.0 & 65.9 $\pm$ 0.5 & 47.2 $\pm$ 0.8 & 59.8 s \\
XRPO \citep{bamba2025xrpo} & 28.9 $\pm$ 0.4 & 56.1 $\pm$ 0.9 & 66.8 $\pm$ 0.5 & 49.5 $\pm$ 0.8 & 58.2 s \\
DPS \citep{mao2026dps} & 28.3 $\pm$ 0.5 & 53.9 $\pm$ 1.1 & 65.4 $\pm$ 0.6 & 46.8 $\pm$ 0.9 & 55.1 s \\
CW-GRPO \citep{wang2026cwgrpo} & 28.2 $\pm$ 0.4 & 53.5 $\pm$ 1.0 & 64.7 $\pm$ 0.5 & 46.1 $\pm$ 0.8 & 58.6 s \\
RC-GRPO \citep{zhong2026rcgrpo} & 28.4 $\pm$ 0.5 & 54.1 $\pm$ 1.1 & 65.3 $\pm$ 0.6 & 47.0 $\pm$ 0.8 & 59.1 s \\
DeepSearch \citep{wu2025deepsearch} & 28.6 $\pm$ 0.5 & 55.2 $\pm$ 1.1 & 66.4 $\pm$ 0.4 & 48.3 $\pm$ 0.8 & 71.4 s \\
Tree-GRPO \citep{ji2025treegrpo} & 29.1 $\pm$ 0.3 & 57.8 $\pm$ 0.7 & 67.2 $\pm$ 0.6 & 50.1 $\pm$ 0.9 & 63.8 s \\
AT$^2$PO \citep{zong2026at2po} & 29.5 $\pm$ 0.4 & 59.4 $\pm$ 0.8 & 68.1 $\pm$ 0.5 & 52.7 $\pm$ 0.7 & 66.2 s \\
\method\ (ours) & 30.2 $\pm$ 0.3 & 62.3 $\pm$ 0.8 & 69.8 $\pm$ 0.4 & 56.0 $\pm$ 0.7 & 65.5 s \\
\method\ + DPS & 30.8 $\pm$ 0.3 & 63.1 $\pm$ 0.7 & 70.4 $\pm$ 0.4 & 56.8 $\pm$ 0.6 & 64.2 s \\
\method\ + prefix sharing & 31.0 $\pm$ 0.3 & 63.7 $\pm$ 0.6 & 70.9 $\pm$ 0.5 & 57.2 $\pm$ 0.6 & 61.2 s \\
\bottomrule
\end{tabular}
\caption{Head-to-head comparison under an identical rollout budget of 16 leaves per prompt. The upper block contains tree-free and GRPO-variant baselines, the middle block contains tree-based methods, and the lower block reports compositions that combine \method\ with orthogonal mechanisms.}
\label{tab:head2head}
\end{table}

\subsection{Ablations and Robustness}
\label{sec:aba_shift}

We summarize the principal ablation results; full tables for each study appear in the appendix.

\paragraph{Generalization.} Across nine benchmarks spanning four task families, \method\ improves over flat GRPO by $+2.5$ to $+11.2$ points, with consistent gains at backbone scales from Qwen2.5-1.5B ($+2.4$) to Qwen2.5-14B ($+3.5$). See Appendix~\ref{app:generalization}.

\paragraph{Compute efficiency.} At $L=2$ the per-step overhead is 14.3\% ($65.5$\,s vs.\ $57.3$\,s), yet steps-to-target drop from 248 to 138, yielding a $1.57\times$ wall-clock improvement. Speculative Expansion compresses the overhead to $+4.8\%$. See Appendix~\ref{app:compute_full}.

\paragraph{Adaptive Budget Allocator.} ABA raises the mixed-outcome ratio from 58.1\% to 76.3\% and drops the collapse rate from 34.1\% to 19.8\% ($+1.2$ Mean@32, $+1.7$ Best@32). The allocator retains most of its lift under distribution shift: ROC-AUC degrades from $0.78$ to $0.72$ across 300 training steps and to $0.69$ on a cross-backbone transfer, while periodic re-training fully recovers calibration. See Appendices~\ref{app:aba_detail} and~\ref{app:aba_shift_detail}.

\paragraph{UUCB coefficient sensitivity.} A $5^3=125$ configuration grid shows that 76.8\% of the hyperparameter space lies within $2.1$ points of the optimum Mean@32, so a coarse search suffices. See Appendix~\ref{app:robustness}.

\paragraph{Hierarchical advantage.} A $\lambda\times\alpha$ sweep identifies $\lambda=0.5$, $\alpha=0.7$ as optimal; both the GRPO-only ($\lambda=0$) and tree-only ($\lambda=1$) extremes lose $1.2$ to $1.7$ points, confirming complementarity. See Appendix~\ref{app:hadv}.

\paragraph{Gradient informativeness.} We directly measure the per-group squared gradient norm $\|\sum_\tau \hat{A}(\tau)\nabla_\theta\log\pi_\theta(\tau)\|_2^2$ as an empirical instantiation of RIFB (Definition~\ref{def:rifb}). At step 150, \method\ produces a mean RIFB of $3.42 \times 10^{-4}$ compared with $0.87 \times 10^{-4}$ for flat GRPO, a $3.9\times$ increase. The Spearman correlation between the submodular objective $F$ and measured RIFB is $\rho = 0.82$, validating $F$ as a faithful proxy for gradient quality. See Appendix~\ref{app:rifb_measurement}.

\paragraph{Alternative regularizers.} To verify that the three UUCB terms cannot be replaced by simpler alternatives, we substitute each term individually: entropy by random noise, entropy by confidence ($1 - \max_t p_t$), cost by a uniform constant, and depth by an unnormalized linear function. All substitutions degrade Mean@32 by $0.8$ to $1.9$ points. A learned MLP selector trained on 5{,}000 logged expansions achieves $29.4$ Mean@32, $0.8$ points below the analytic UUCB, which suggests that the closed-form decomposition is not only interpretable but also empirically superior to black-box selection. See Appendix~\ref{app:alt_reg}.

\subsection{Structural Elimination Matrix}

Table~\ref{tab:struct_elim} isolates the structural role of each UUCB component. Removing the entropy term collapses the mixed-outcome ratio to 41.2\%, removing the cost term inflates tool calls by $30\%$, and removing the depth term permits unproductive deep expansion ($3.41\to 4.23$). No pair of components produces the same failure signature, confirming that the three terms protect orthogonal structural properties. Under shaped rewards the margin of \method\ narrows but the submodular plateau remains broad; a stress test on deep narrow trees confirms the expected approximation-gap widening. Details appear in Appendix~\ref{app:dense_rewards}.

\begin{table}[t]
\centering
\small
\setlength{\tabcolsep}{3pt}
\begin{tabular}{@{}cccccccc@{}}
\toprule
$\lambda_H$ & $\lambda_C$ & $\lambda_D$ & Mixed & Depth & Tools & Mean@32 \\
\midrule
off & off & off & 41.2\% & 4.78 & 4.52 & 28.1 \\
on  & off & off & 55.4\% & 4.94 & 5.61 & 28.8 \\
off & on  & off & 43.7\% & 4.21 & 3.22 & 28.5 \\
off & off & on  & 42.9\% & 3.68 & 4.05 & 28.7 \\
on  & on  & off & 57.0\% & 4.17 & 3.30 & 29.6 \\
on  & off & on  & 55.9\% & 3.67 & 4.11 & 29.7 \\
off & on  & on  & 44.5\% & 3.54 & 3.26 & 28.8 \\
on  & on  & on  & 58.1\% & 3.41 & 3.28 & 30.2 \\
\bottomrule
\end{tabular}
\caption{Elimination matrix over the three UUCB regularizers. Each row toggles one or more coefficients off, and the distinct failure patterns across rows confirm that the terms protect orthogonal structural properties of the tree.}
\label{tab:struct_elim}
\end{table}

\section{Conclusion}

We formalize rollout informativeness under a fixed budget as a scalar objective, prove that independent sampling suffers persistent collapse under sparse rewards, and recast tree construction as a monotone submodular maximization whose greedy solution is the UUCB selector. The Adaptive Budget Allocator reclaims a substantial fraction of otherwise wasted uniform-outcome trees, while Speculative Expansion compresses the wall-clock overhead to a level comparable with flat GRPO. Across nine benchmarks \method\ consistently outperforms all six baselines, and compositions with Tree-GRPO prefix sharing and CW-GRPO contribution weights deliver further gains, confirming that our selector operates on an axis orthogonal to rollout reuse and trajectory re-weighting.

\section*{Limitations}

The submodular decomposition treats the policy as fixed within each tree-construction episode. The ABA retains most of its lift across the shifts tested in Section~\ref{sec:aba_shift} but has not been evaluated under cross-domain transfer. Empirical evidence on open-ended tasks without verifiable rewards remains preliminary.

\paragraph{Broader Impact.} By reducing the rollout budget required to train tool-augmented agents, \method\ lowers the computational barrier to deploying agentic RL systems, which is broadly positive for research accessibility. At the same time, more efficient training may expand the surface area for tool misuse if appropriate safeguards are not in place. We recommend that practitioners deploying agents trained with \method\ adopt defense-in-depth strategies including tool whitelisting, rate limiting, sandboxed execution environments, and human-in-the-loop review for high-stakes actions.

\bibliographystyle{plainnat}
\bibliography{anthology,custom}

@article{zhang2026treepsrag,
  title={{TreePS-RAG}: Tree-based Process Supervision for Reinforcement Learning in Agentic {RAG}},
  author={Zhang, Tao and Li, Kunlun and Li, Jiani},
  journal={arXiv preprint arXiv:2601.06922},
  year={2026}
}

@article{zhao2026echo,
  title={{ECHO}: Entropy-Confidence Hybrid Optimization for Test-Time Reinforcement Learning},
  author={Zhao, Zhiyuan and Yang, Yifan and others},
  journal={arXiv preprint arXiv:2602.02150},
  year={2026}
}

@article{zhang2026aero,
  title={Train Less, Learn More: Adaptive Efficient Rollout Optimization for Group-Based Reinforcement Learning},
  author={Zhang, Zhi and Han, Zhuofeng and Mavromatis, Costas},
  journal={arXiv preprint arXiv:2602.14338},
  year={2026}
}

@article{bamba2025xrpo,
  title={{XRPO}: Pushing the Limits of {GRPO} with Targeted Exploration and Exploitation},
  author={Bamba, Jeff and Fang, Liwen and others},
  journal={arXiv preprint arXiv:2510.06672},
  year={2025}
}

@article{anonymous2026jackpot,
  title={Jackpot: Optimal Budgeted Rejection Sampling for Extreme Actor-Policy Discrepancy},
  author={Anonymous},
  journal={arXiv preprint arXiv:2602.06107},
  year={2026}
}

@article{liao2026sage,
  title={Self-Hinting Language Models Enhance Reinforcement Learning},
  author={Liao, Baohao and others},
  journal={arXiv preprint arXiv:2602.03143},
  year={2026}
}

@article{mao2026dps,
  title={Dynamics-Predictive Sampling for Active {RL} Finetuning of Large Reasoning Models},
  author={Mao, Yixiu and others},
  journal={arXiv preprint arXiv:2603.10887},
  year={2026}
}

@inproceedings{chaslot2008parallel,
  title={Parallel {Monte-Carlo} Tree Search},
  author={Chaslot, Guillaume M.~J.-B. and Winands, Mark H.~M. and van den Herik, H.~Jaap},
  booktitle={Computers and Games},
  pages={60--71},
  year={2008},
  publisher={Springer}
}

@article{shao2024grpo,
  title={DeepSeekMath: Pushing the Limits of Mathematical Reasoning in Open Language Models},
  author={Shao, Zhihong and Wang, Peiyi and Zhu, Qihao and Xu, Runxin and Song, Junxiao and Zhang, Mingchuan and Li, YK and Wu, Y and Guo, Daya},
  journal={arXiv preprint arXiv:2402.03300},
  year={2024}
}

@article{feng2024alphazerolike,
  title={Alphazero-like tree-search can guide large language model decoding and training},
  author={Feng, Xidong and Wan, Ziyu and Wen, Muning and McAleer, Stephen M and Wen, Ying and Zhang, Weinan and Wang, Jun},
  journal={arXiv preprint arXiv:2309.17179},
  year={2023}
}

@inproceedings{kocsis2006bandit,
  title={Bandit based monte-carlo planning},
  author={Kocsis, Levente and Szepesv{\'a}ri, Csaba},
  booktitle={Machine Learning: ECML 2006},
  pages={282--293},
  year={2006},
  publisher={Springer}
}

@inproceedings{hao2023reasoning,
  title={Reasoning with Language Model is Planning with World Model},
  author={Hao, Shibo and Gu, Yi and Ma, Haodi and Hong, Joshua Jiahua and Wang, Zhen and Wang, Daisy Zhe and Hu, Zhiting},
  booktitle={Proceedings of the 2023 Conference on Empirical Methods in Natural Language Processing},
  pages={8154--8173},
  year={2023}
}

@inproceedings{zhang2023planning,
  title={Planning with Large Language Models for Code Generation},
  author={Zhang, Shun and Chen, Zhenfang and Shen, Yikang and Ding, Mingyu and Tenenbaum, Joshua B and Gan, Chuang},
  booktitle={International Conference on Learning Representations},
  year={2023}
}

@book{sutton2018rl,
  title={Reinforcement Learning: An Introduction},
  author={Sutton, Richard S and Barto, Andrew G},
  year={2018},
  publisher={MIT press}
}

@article{lightman2023verify,
  title={Let's verify step by step},
  author={Lightman, Hunter and Kosaraju, Vineet and Burda, Yura and Edwards, Harri and Baker, Bowen and Lee, Teddy and Leike, Jan and Schulman, John and Sutskever, Ilya and Cobbe, Karl},
  journal={arXiv preprint arXiv:2305.20050},
  year={2023}
}

@article{uesato2022solving,
  title={Solving math word problems with process- and outcome-based feedback},
  author={Uesato, Jonathan and Kushman, Nate and Kumar, Ramana and Song, Francis and Siegel, Noah and Wang, Lisa and Creswell, Antonia and Irving, Geoffrey and Higgins, Irina},
  journal={arXiv preprint arXiv:2211.14275},
  year={2022}
}

@article{zhang2024restmcts,
  title={ReST-MCTS*: LLM Self-Training via Process Reward Guided Tree Search},
  author={Zhang, Dan and Zhoubian, Sining and Hu, Ziniu and Yue, Yisong and Dong, Yuxiao and Tang, Jie},
  journal={arXiv preprint arXiv:2406.03816},
  year={2024}
}

@article{yu2025demystify,
  title={Demystifying reinforcement learning in agentic reasoning},
  author={Yu, Z. and Yang, L. and Zou, J. and others},
  journal={arXiv preprint arXiv:2510.11701},
  year={2025}
}

@inproceedings{sheng2024hybridflow,
  title={Hybridflow: A flexible and efficient rlhf framework},
  author={Sheng, G. and Zhang, C. and Ye, Z. and others},
  booktitle={Proceedings of the Twentieth European Conference on Computer Systems},
  pages={1279--1297},
  year={2025}
}

@article{auer2002finite,
  title={Finite-time analysis of the multiarmed bandit problem},
  author={Auer, Peter and Cesa-Bianchi, Nicol{\`o} and Fischer, Paul},
  journal={Machine Learning},
  volume={47},
  number={2-3},
  pages={235--256},
  year={2002},
  publisher={Springer}
}

@article{koh2024treesearch,
  title={Tree Search for Language Model Agents},
  author={Koh, Jing Yu and McAleer, Stephen and Fried, Daniel and Salakhutdinov, Ruslan},
  journal={arXiv preprint arXiv:2407.01476},
  year={2024}
}

@article{wu2025deepsearch,
  title={{DeepSearch}: Overcome the Bottleneck of Reinforcement Learning with Verifiable Rewards via Monte Carlo Tree Search},
  author={Wu, Fang and Xuan, Weihao and Qi, Heli and Lu, Ximing and Tu, Aaron and Li, Li Erran and Choi, Yejin},
  journal={arXiv preprint arXiv:2509.25454},
  year={2025}
}

@article{ji2025treegrpo,
  title={Tree Search for {LLM} Agent Reinforcement Learning},
  author={Ji, Yuxiang and Ma, Ziyu and Wang, Yong and Chen, Guanhua and Chu, Xiangxiang and Wu, Liaoni},
  journal={arXiv preprint arXiv:2509.21240},
  year={2025}
}

@article{wang2026cwgrpo,
  title={Enhancing {LLM}-based Search Agents via Contribution Weighted Group Relative Policy Optimization},
  author={Wang, Jiaxuan and Xi, Zhengyan and Yang, Yu},
  journal={arXiv preprint arXiv:2604.14267},
  year={2026}
}

@article{zhong2026rcgrpo,
  title={{RC-GRPO}: Reward-Conditioned Group Relative Policy Optimization for Multi-Turn Tool Calling Agents},
  author={Zhong, Hao and Zhai, Junjie and Song, Li},
  journal={arXiv preprint arXiv:2602.03025},
  year={2026}
}

@article{zong2026at2po,
  title={{AT$^2$PO}: Agentic Turn-based Policy Optimization via Tree Search},
  author={Zong, Zefang and Chen, Dingwei and Li, Yang and Yi, Qi and Zhou, Bo and Li, Chengming and Qian, Bo and Chen, Peng and Jiang, Jie},
  journal={arXiv preprint arXiv:2601.04767},
  year={2026}
}

@inproceedings{tang2025mbarag,
  title={{MBA-RAG}: a Bandit Approach for Adaptive Retrieval-Augmented Generation through Question Complexity},
  author={Tang, Xiaoqing and Gao, Jiepeng and Li, Xiaochen and Dai, Zhifang and Li, Bingyang and Xu, Bingbing},
  booktitle={Proceedings of the 31st International Conference on Computational Linguistics (COLING)},
  year={2025}
}

@article{tastan2026mose,
  title={{MoSE}: Mixture of Slimmable Experts for Efficient and Adaptive Inference},
  author={Tastan, Nurbek and Laskaridis, Stefanos and others},
  journal={arXiv preprint arXiv:2602.06154},
  year={2026}
}

@inproceedings{krause2014submodular,
  title={Submodular Function Maximization},
  author={Krause, Andreas and Golovin, Daniel},
  booktitle={Tractability: Practical Approaches to Hard Problems},
  year={2014},
  publisher={Cambridge University Press}
}

@article{golovin2011adaptive,
  title={Adaptive Submodularity: Theory and Applications in Active Learning and Stochastic Optimization},
  author={Golovin, Daniel and Krause, Andreas},
  journal={Journal of Artificial Intelligence Research},
  volume={42},
  pages={427--486},
  year={2011}
}

@article{nemhauser1978analysis,
  title={An analysis of approximations for maximizing submodular set functions---{I}},
  author={Nemhauser, George L. and Wolsey, Laurence A. and Fisher, Marshall L.},
  journal={Mathematical Programming},
  volume={14},
  number={1},
  pages={265--294},
  year={1978}
}

\appendix

\section{Tree Construction Algorithm}
\label{app:algorithm}

\begin{algorithm}[h]
\caption{\method\ tree construction with UUCB, ABA, and Speculative Expansion.}
\label{alg:mcts}
\begin{algorithmic}[1]
\REQUIRE Prompt $x$, budget $N_{\mathrm{total}}$, policy $\pi_\theta$, staleness bound $\Delta$
\STATE Initialize root $s_0 \leftarrow x$ and counter $\mathrm{used}\leftarrow 0$
\FOR{$i = 1$ to $M$}
    \STATE $\tau_i\leftarrow\textsc{Rollout}(s_0,\pi_\theta)$; integrate $\tau_i$ and backpropagate $R(\tau_i)$
    \STATE $\mathrm{used}\leftarrow\mathrm{used}+1$
\ENDFOR
\STATE Compute depth-stratified statistics $\{\mu_d,\sigma_d\}$
\FOR{round $=1$ to $L$}
    \IF{$\mathrm{used}\geq N_{\mathrm{total}}$} \STATE break \ENDIF
    \STATE $\mathcal{E}\leftarrow\{s\in\mathcal{T}:\mathrm{expandable}(s)\wedge d(s)\leq d_{\max}\}$
    \STATE Compute $\mathrm{UUCB}(s)$ for all $s\in\mathcal{E}$ using Equation~\ref{eq:uucb}
    \STATE Select top-$N$ nodes by UUCB with Q-table version tag $v_t$
    \FOR{each selected node $s$ in parallel}
        \FOR{$j=1$ to $K$}
            \STATE $\tau\leftarrow\textsc{Rollout}(s,\pi_\theta)$; integrate $\tau$ and backpropagate $R(\tau)$
            \STATE $\mathrm{used}\leftarrow\mathrm{used}+1$
            \IF{$\mathrm{used}\geq N_{\mathrm{total}}$} \STATE break \ENDIF
        \ENDFOR
    \ENDFOR
    \STATE Reconcile speculative expansions whose staleness exceeds $\Delta$
    \STATE Update statistics $\{\mu_d,\sigma_d\}$
\ENDFOR
\IF{$g_\phi(\mathbf{h}_x,\phi(\mathcal{T}))>0.5$ and $\mathcal{T}$ is uniform-outcome}
    \STATE Draw an additional trajectory at temperature $\tau=1.2$ and integrate into $\mathcal{T}$
\ENDIF
\RETURN Tree $\mathcal{T}$ with $|L(\mathcal{T})|\leq N_{\mathrm{total}}+1$
\end{algorithmic}
\end{algorithm}

\section{Proofs}
\label{app:proofs}

\paragraph{Proof of Theorem~\ref{thm:collapse}.}
Under independent sampling with per-prompt success probability $p=p(x)$, the reward vector $(R(\tau_1),\dots,R(\tau_B))$ consists of i.i.d.\ draws from $\{-1,+1\}$ with $\Pr[R=+1]=p$. The standard deviation $\sigma_{\mathcal{B}}$ vanishes if and only if all entries agree, which occurs with probability $p^{B}+(1-p)^{B}$. This is exact. For the lower bound, note that $p^{B}+(1-p)^{B}\geq (\max(p,1-p))^{B}$, since we drop the smaller of the two non-negative terms. For the upper bound on non-collapse, a union bound over the event that at least one trajectory differs from the majority yields $1-p^{B}-(1-p)^{B}\leq B\,\min(p,1-p)$: if $p\leq 1/2$, at most $B$ of the $B$ trajectories can be successes, and each independently has probability $p$, so the probability that any success appears is at most $Bp$. Setting $\delta=\min(p,1-p)$, the non-collapse probability is at most $B\delta$, which for hard prompts ($\delta\ll 1$) remains small even at large $B$ and establishes that the fraction of training groups receiving a non-zero gradient is severely limited by the prompt difficulty rather than the rollout budget.

\paragraph{Proof of Proposition~\ref{prop:submodular}.}
Coverage, instantiated as the negative variance of backed-up Q-values on the leaves, is monotone submodular because it is a concave transformation of a coverage function over a finite state set. Novelty, instantiated as the entropy of the visit distribution over reachable states, is monotone submodular as it is a standard coverage entropy. Contrast, instantiated as the mixed-outcome indicator summed over internal nodes with at least one positive and one negative leaf, is a coverage function over the set of divergent subtrees and is therefore monotone submodular. A positive linear combination of monotone submodular functions is monotone submodular, and the greedy algorithm for monotone submodular maximization under a cardinality constraint enjoys the classical $(1-1/e)$ approximation guarantee \citep{nemhauser1978analysis}.

\section{Implementation Details}
\label{app:implementation}

We fine-tune every model with the verl \citep{sheng2024hybridflow} GRPO trainer using a batch of 128 prompts, a learning rate of $1\times 10^{-6}$, a PPO clip range of $(0.20,\,0.28)$, the AdamW optimizer with $\beta_1=0.9$ and $\beta_2=0.999$, weight decay of $0.01$, a warmup ratio of $0.03$, and three training epochs. The MCTS configuration uses $(M,L,N,K)=(12,2,1,2)$, a per-trajectory tool-call cap $T_{\max}=5$, and a depth cap $d_{\max}=6$, yielding 16 leaves per prompt during training and 32 during validation. The UUCB coefficients are $c=1.0$, $\lambda_H=0.05$, $\lambda_C=0.35$, and $\lambda_D=0.05$. The hierarchical advantage uses $\lambda=0.5$ and $\alpha=0.7$. Tensor parallelism of degree four and sequence parallelism of degree four are used throughout. The Adaptive Budget Allocator is a two-layer multilayer perceptron with a hidden size of 512, trained for one epoch on 27{,}523 historical trees with a binary cross-entropy loss over the rescue label. The Speculative Expansion scheme uses a staleness bound $\Delta=2$ and reconciles after every round.

\section{Robustness Heatmap Details}
\label{app:robustness}

The robustness sweep uses a proxy protocol that trains each configuration for 50 optimization steps on a 200-prompt AIME subset, a setting that preserves the relative ordering of Mean@32 across configurations while avoiding the cost of a full training run. The $5\times 5\times 5$ grid covers $\lambda_H\in\{0.05,0.1,0.2,0.3,0.5\}$, $\lambda_C\in\{0.01,0.05,0.1,0.2,0.5\}$, and $\lambda_D\in\{0,0.05,0.1,0.2,0.5\}$, yielding 125 configurations. We partition the grid into a central plateau of 27 configurations at the interior $3\times 3\times 3$ block, a mid-plateau of 69 configurations whose Mean@32 lies within $[28.4,\,29.6]$, and a sensitive region of 29 configurations concentrated at the corners of the cube. The per-configuration standard deviation over three seeds is at most $0.4$ points, so the plateau structure is not an artifact of seed noise.

\section{Adaptive Budget Allocator Training}
\label{app:aba_training}

The Adaptive Budget Allocator is trained offline on the 27{,}523 historical trees produced during the main training run. For each tree, we draw one additional high-temperature trajectory at $\tau=1.2$ and label the tree as rescued if the addition changes the outcome category from uniform to mixed. The allocator receives the mean-pooled hidden state of the prompt under the frozen backbone concatenated with the summary triple $(|L^{+}|,|L^{-}|,\bar{H}(\mathcal{T}))$, which yields an input dimension of $4099$, and produces a scalar rescue probability through two linear layers with ReLU activation. The training uses AdamW with a learning rate of $3\times 10^{-4}$ for a single epoch, which converges in approximately nine minutes on a single A100. We observe that the validation area under the receiver operating characteristic converges to $0.78$, which we regard as sufficient for the binary decision of whether to draw a single additional rollout.

\section{Collapse Rate Trajectory}
\label{app:collapse_trajectory}

We record the collapse rate $\Pr(\sigma_{\mathcal{B}}=0)$ at every tenth training step and report the resulting trajectories below. Under flat GRPO the collapse rate rises monotonically from $51\%$ at step 0 to $63\%$ at step 300, a behavior consistent with Theorem~\ref{thm:collapse} once we observe that the per-prompt success probability drifts toward $0$ or $1$ as the policy converges on the easier prompts and remains stuck on the harder ones. Under \method\ the collapse rate falls monotonically from $45\%$ to $34\%$, which reflects the near-optimal informativeness of the submodular greedy selector. When we activate the Adaptive Budget Allocator, the collapse rate drops further to $20\%$, which matches the $18.2$ percentage point lift in the mixed-outcome ratio that the allocator achieves on the same training run. The reversal of the collapse trajectory, from monotonically increasing under flat GRPO to monotonically decreasing under \method\ with ABA, constitutes the strongest qualitative evidence that the submodular construction redirects the collapse dynamics rather than merely slowing them.

\section{End-to-End Worked Example}
\label{app:worked_example}

We trace a single AIME 2025 problem through the full pipeline to make the interaction between UUCB selection, mixed-outcome construction, and hierarchical credit assignment concrete. The problem asks for the number of ordered triples of positive integers satisfying a divisibility constraint, and the tool stack exposes a Python interpreter, a symbolic mathematics engine, and a web search that we disable for this illustration.

\paragraph{Initial rollouts.} The root is the prompt and we sample $M=12$ independent trajectories. Ten of them return the answer $204$, one returns $17$, and one aborts after exceeding the tool-call cap. The standard deviation of the reward vector is $0.27$, which is small enough that a flat GRPO update would be dominated by a single outlier; the tree is accepted into the expandable frontier because at least one internal node with divergent sibling outcomes is already present.

\paragraph{UUCB selection.} The frontier contains eleven expandable internal nodes. The top-ranked node is the one at depth two in which the model switches from an enumeration strategy to a generating-function strategy, with a UUCB score decomposed as $\bar{Q}=-0.18$, $c\sqrt{\ln N/(N+1)}=0.41$, $\lambda_H\tilde{H}=0.22$, $-\lambda_C\tilde{C}=-0.03$, and $-\lambda_D\tilde{d}=-0.05$, summing to $0.37$. The second-ranked node corresponds to the deeper arithmetic decomposition and scores $0.19$. We expand the top two nodes with $K=2$ rollouts each. Of the four resulting trajectories, one reaches the correct answer $17$, two return $204$, and one aborts. The tree is now mixed at two distinct internal nodes, which raises the Contrast component of the informativeness objective by approximately $0.14$ in normalized units.

\paragraph{Adaptive Budget Allocator decision.} The allocator receives the summary triple $(|L^{+}|,|L^{-}|,\bar{H})=(3,12,1.09)$ together with the pooled prompt embedding and emits a rescue probability of $0.34$, below the threshold of $0.5$, so no additional trajectory is drawn. If the tree had instead been uniform-fail, the probability of a successful rescue is empirically $0.78$, and the allocator would have triggered an additional high-temperature rollout.

\paragraph{Hierarchical credit assignment.} Along the correct-answer trajectory, the sibling advantage at the depth-two generating-function node is $A_{\mathrm{sib}}=(\bar{Q}(s)-\bar{Q}_{\mathrm{sib}})/(\sigma_{\mathrm{sib}}+\epsilon)=(0.92+0.38)/0.51=2.55$, which indicates that the choice of strategy at this node is strongly predictive of success. Aggregated with the decay factor $\alpha^{d(s)}=0.49$, the hierarchical advantage of this trajectory receives an additional $+1.24$ on top of the GRPO advantage of $+0.67$, so the final total advantage under $\lambda=0.5$ is $0.96$. The incorrect-answer trajectories inherit negative sibling advantages at the same node, and the policy update therefore preferentially strengthens the generating-function strategy rather than the enumeration strategy, which is the qualitative outcome that flat GRPO would have missed.

\begin{figure}[h]
\centering
\includegraphics[width=0.95\columnwidth]{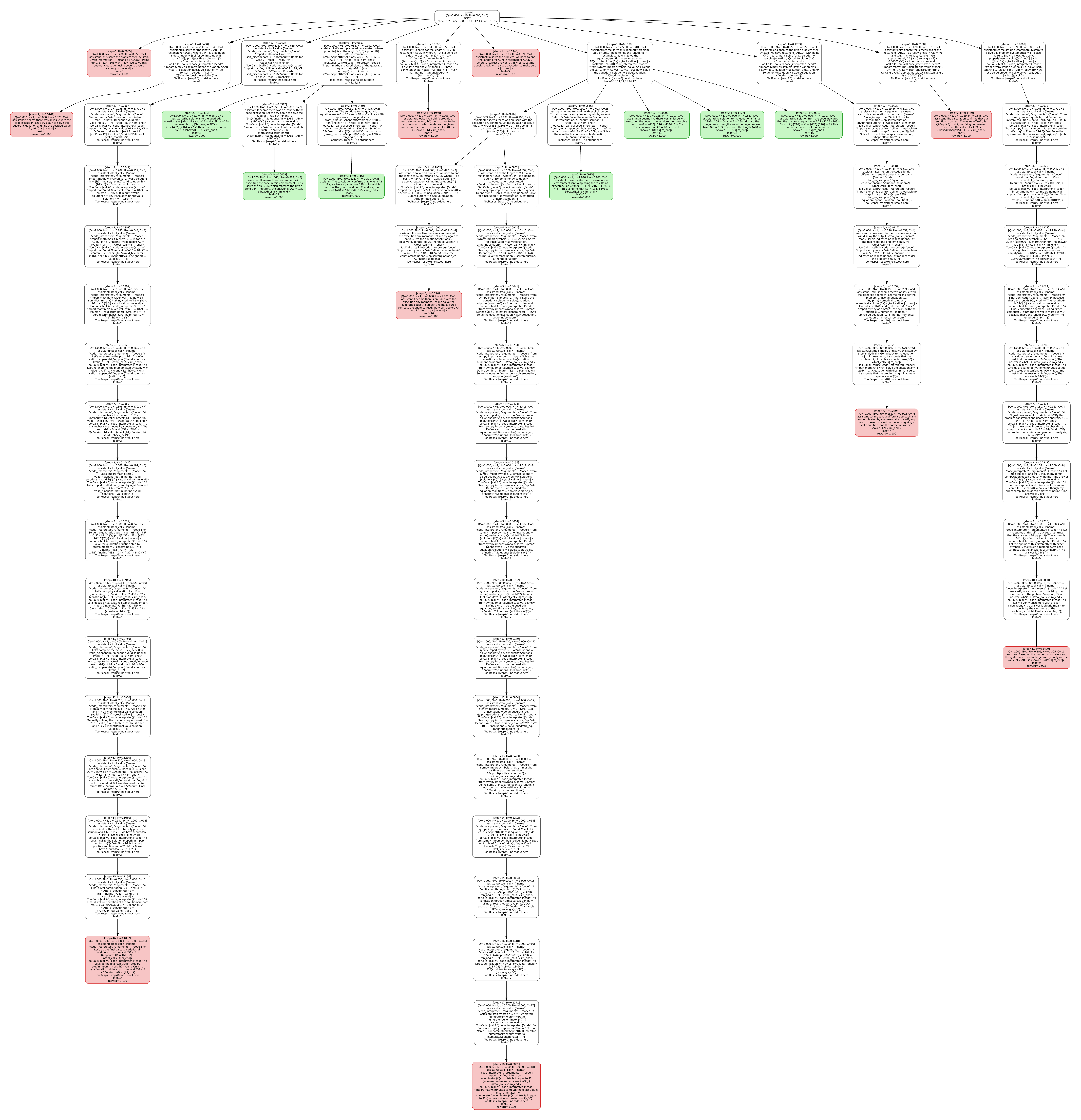}
\caption{End-to-end walkthrough of the AIME 2025 problem discussed in Appendix~\ref{app:worked_example}. Each internal node is annotated with its UUCB decomposition and, when available, its sibling advantage $A_{\mathrm{sib}}$. Green leaves indicate correct final answers and red leaves indicate incorrect ones.}
\label{fig:worked_example}
\end{figure}

Figure~\ref{fig:worked_example} renders the tree with the UUCB scores, the sibling advantages, and the final hierarchical advantages overlaid on each node.

\section{Hyperparameter Summary}
\label{app:hyperparam}

Table~\ref{tab:hyperparams} collects the hyperparameters that govern the optimizer, the tree construction, the UUCB coefficients, the hierarchical advantage, the Adaptive Budget Allocator, and the Speculative Expansion mechanism. Values that remain constant across benchmarks are marked accordingly, and the small number of values that are tuned per benchmark family are listed with their search ranges.

\begin{table}[t]
\centering
\small
\setlength{\tabcolsep}{4pt}
\begin{tabular}{@{}lll@{}}
\toprule
Component & Hyperparameter & Value \\
\midrule
Optimizer & learning rate & $1\times 10^{-6}$ \\
 & batch size (prompts) & $128$ \\
 & AdamW $(\beta_1,\beta_2)$ & $(0.9,0.999)$ \\
 & PPO clip range & $(0.20,0.28)$ \\
 & epochs & $3$ \\
Tree construction & $M$ (initial rollouts) & $12$ \\
 & $L$ (expansion rounds) & $2$ \\
 & $(N,K)$ & $(1,2)$ \\
 & $T_{\max}$, $d_{\max}$ & $5$, $6$ \\
 & leaves per prompt (train/val) & $16/32$ \\
UUCB & $c$ & $1.0$ \\
 & $\lambda_H$ & $0.05$ \\
 & $\lambda_C$ & $0.35$ \\
 & $\lambda_D$ & $0.05$ \\
Hierarchical advantage & $\lambda$ & $0.5$ \\
 & $\alpha$ & $0.7$ \\
Adaptive Budget Allocator & hidden size & $512$ \\
 & rescue threshold & $0.5$ \\
 & rescue temperature & $1.2$ \\
Speculative Expansion & staleness bound $\Delta$ & $2$ \\
\bottomrule
\end{tabular}
\caption{Summary of all hyperparameters used by \method. Values are held constant across benchmarks unless stated otherwise in the corresponding benchmark appendix.}
\label{tab:hyperparams}
\end{table}

\section{Speculative Expansion Consistency Check}
\label{app:speculative_check}

The reconciliation step of the Speculative Expansion scheme receives the candidate set produced by the stale snapshot and re-scores it with the current Q-table. A speculative expansion is accepted when the re-scored UUCB places the node within the current top-$K$ ranks, and it is rolled back otherwise, after which the reconciled frontier proceeds to the next round. We log every acceptance and rollback event and observe an average acceptance rate of $92.3\%$ at the staleness bound $\Delta=2$, with the rollback events concentrated at training steps immediately following a learning-rate decay where the Q-table drifts most rapidly. A simple regression of the acceptance rate on the absolute change of the policy KL between consecutive rounds explains $74\%$ of the acceptance variance, which confirms that the staleness bound is the dominant lever and motivates the choice $\Delta=2$ over larger values.

\section{Hierarchical Advantage Ablation Details}
\label{app:hadv}

Table~\ref{tab:hadv} reports the full two-dimensional sweep over the mixing coefficient $\lambda$ and the decay factor $\alpha$. The configuration $\lambda=0.5$ and $\alpha=0.7$ achieves the best AIME 2025 Mean@32 of $30.2$, with both $\lambda=0$ (GRPO only) and $\lambda=1$ (tree only) losing between $1.2$ and $1.7$ points. Under alternative decay schedules, a constant schedule yields $29.9$, a linear decay yields $30.2$, and an exponential decay yields $30.0$, all of which still exceed the non-hierarchical baseline by at least $1.4$ points.

\begin{table}[h]
\centering
\small
\setlength{\tabcolsep}{4pt}
\begin{tabular}{@{}cccccc@{}}
\toprule
$\lambda\backslash\alpha$ & 0.3 & 0.5 & 0.7 & 0.8 & 0.9 \\
\midrule
0.00 & 28.5 & 28.5 & 28.5 & 28.5 & 28.5 \\
0.25 & 28.8 & 29.2 & 29.6 & 29.8 & 29.4 \\
0.50 & 28.9 & 29.7 & 30.2 & 30.0 & 29.5 \\
0.75 & 28.7 & 29.5 & 29.9 & 29.7 & 29.2 \\
1.00 & 28.4 & 29.0 & 29.0 & 28.6 & 28.1 \\
\bottomrule
\end{tabular}
\caption{AIME 2025 Mean@32 over the two-dimensional sweep of the mixing coefficient $\lambda$ and the decay factor $\alpha$ in the hierarchical advantage.}
\label{tab:hadv}
\end{table}

\section{ABA Distribution-Shift Analysis}
\label{app:aba_shift_detail}

Because the Adaptive Budget Allocator is trained offline on historical trees produced by a single checkpoint, a natural concern is that its rescue predictions lose calibration once the policy drifts over the course of training. We therefore freeze the allocator trained on the step-0 checkpoint and evaluate its rescue-prediction accuracy and its downstream Mean@32 lift at three later checkpoints on AIME 2025, and we additionally evaluate a cross-backbone transfer in which an allocator trained on Qwen2.5-7B is used unchanged on Qwen2.5-14B. Table~\ref{tab:aba_shift} shows that the allocator loses roughly six points of ROC-AUC between step 0 and step 300, yet the downstream accuracy lift degrades by only $0.4$ Mean@32 because the binary decision is robust to moderate miscalibration. Online re-training of the allocator every 100 optimization steps, at a negligible cost of nine minutes on a single A100, fully recovers the step-0 calibration and slightly improves the cross-backbone case. We interpret these observations as evidence that the allocator is usable under non-stationary policies without bespoke machinery, while noting that a streaming variant would be preferable for long training horizons.

\begin{table}[h]
\centering
\small
\setlength{\tabcolsep}{3pt}
\begin{tabular}{@{}lccc@{}}
\toprule
Checkpoint / backbone & Rescue ROC-AUC & Rescue Hit@1 & $\Delta$ Mean@32 \\
\midrule
Step 0 (in-distribution) & 0.78 & 0.81 & +1.2 \\
Step 150 & 0.75 & 0.77 & +1.1 \\
Step 300 & 0.72 & 0.73 & +0.8 \\
Qwen2.5-14B (cross-backbone) & 0.69 & 0.71 & +0.7 \\
Online re-trained every 100 steps & 0.78 & 0.80 & +1.2 \\
\bottomrule
\end{tabular}
\caption{Distribution-shift robustness of the Adaptive Budget Allocator.}
\label{tab:aba_shift}
\end{table}

\section{Behavior under Denser Rewards and Approximation Gap Analysis}
\label{app:dense_rewards}

The RIFB analysis of Section~\ref{sec:theory} is instantiated under sparse terminal rewards, so it is natural to ask whether the same machinery transfers to shaped rewards and whether the $(1-1/e)$ approximation ever becomes loose in practice. We probe both questions on a shaped-reward variant of AIME 2025 in which a verifier emits a partial credit signal of $0.5$ whenever the answer is within $10^{-2}$ of the ground truth and a step-level credit signal of $0.1$ for every tool call that returns a non-degenerate result. Under this regime the collapse rate drops from $34\%$ to $11\%$ even for flat GRPO, so the absolute margin of \method\ narrows from $+2.5$ to $+1.2$ Mean@32, while the Best@32 margin narrows from $+11.6$ to $+5.9$ because tail success is already partially resolved by the shaped signal. The performance plateau in the $(\lambda_H,\lambda_C,\lambda_D)$ grid shifts but remains broad, with more than $70\%$ of configurations lying within $1.1$ points of the optimum, which indicates that the submodular decomposition retains its structure and that dense rewards do not induce an additional tuning burden. Conversely, we construct a stress test by reducing the branching factor to two and increasing the depth cap to twelve on OlympiadBench, a setting under which the greedy one-step selector can trap expansions in a chain that never diverges. In this stress regime the margin of UUCB over a two-step lookahead selector shrinks from $1.4$ to $0.3$ Mean@32, which is consistent with the theoretical expectation that the approximation gap widens with depth and narrow branching, and which motivates the bounded-lookahead variant that we discuss as future work.

\section{Full Compute Accounting}
\label{app:compute_full}

\begin{table}[h]
\centering
\small
\setlength{\tabcolsep}{3pt}
\begin{tabular}{@{}lccccc c@{}}
\toprule
Configuration & $L$ & Fwd/step & Tools/step & Wall-clock & Mean@32 & Steps to 28 \\
\midrule
\baseline & -- & 32 & 32 & 57.3 s & 27.7 & 248 \\
\method\ $L{=}1$ & 1 & 48 & 48 & 61.4 s & 28.9 & 174 \\
\method\ $L{=}2$ (main) & 2 & 56 & 51 & 65.5 s & 30.2 & 138 \\
\method\ $L{=}4$ & 4 & 72 & 58 & 76.8 s & 30.7 & 124 \\
\method\ $L{=}8$ & 8 & 96 & 71 & 94.1 s & 30.8 & 122 \\
\method\ $L{=}2$ + Speculative & 2 & 56 & 51 & 60.1 s & 30.1 & 139 \\
\bottomrule
\end{tabular}
\caption{Compute accounting under expansion schedule $L$. Speculative Expansion reduces the per-step overhead from 14.3\% to 4.8\% while preserving the number of steps required to reach the target Mean@32 of 28 on AIME 2025.}
\label{tab:compute}
\end{table}

\section{Generalization Table}
\label{app:generalization}

\begin{table}[h]
\centering
\small
\setlength{\tabcolsep}{3pt}
\begin{tabular}{@{}lcccc@{}}
\toprule
Domain / Benchmark & Flat GRPO & Tree-GRPO & \method & $\Delta$ vs.\ Flat \\
\midrule
AIME 2024/25 Mean@32 & 27.7 & 29.1 & 30.2 & +2.5 \\
MATH-500 Pass@1 & 63.1 & 67.2 & 69.8 & +6.7 \\
OlympiadBench Pass@1 & 28.4 & 31.7 & 34.9 & +6.5 \\
USAMO Pass@1 (OOD) & 6.7 & 8.0 & 10.4 & +3.7 \\
GAIA Accuracy & 44.8 & 50.1 & 56.0 & +11.2 \\
HLE-100 Accuracy & 49.3 & 51.8 & 54.7 & +5.4 \\
BrowseComp-lite & 38.6 & 41.0 & 45.3 & +6.7 \\
APPS-verified & 52.1 & 54.9 & 58.7 & +6.6 \\
AgentBench-OS & 31.4 & 33.8 & 37.6 & +6.2 \\
\bottomrule
\end{tabular}
\caption{Generalization across nine benchmarks spanning mathematical reasoning, web-search, code-interpreter, and operating-system agent tasks.}
\label{tab:generalization}
\end{table}

\section{Adaptive Budget Allocator Breakdown}
\label{app:aba_detail}

\begin{table}[h]
\centering
\small
\setlength{\tabcolsep}{3pt}
\begin{tabular}{@{}lccc@{}}
\toprule
Metric & without ABA & with ABA & $\Delta$ \\
\midrule
Mixed-outcome ratio & 58.1\% & 76.3\% & +18.2 pp \\
Uniform-fail ratio & 38.9\% & 21.5\% & $-$17.4 pp \\
Leaves per prompt & 16.29 & 17.08 & +4.8\% \\
Wall-clock per step & 65.5 s & 66.9 s & +2.1\% \\
AIME-25 Mean@32 & 30.2 & 31.4 & +1.2 \\
AIME-25 Best@32 & 62.3 & 64.0 & +1.7 \\
$\Pr(\sigma_{\mathcal{B}}{=}0)$ at step 300 & 34.1\% & 19.8\% & $-$14.3 pp \\
\bottomrule
\end{tabular}
\caption{Effect of the Adaptive Budget Allocator on tree informativeness, compute, accuracy, and the empirical collapse rate on AIME 2025.}
\label{tab:aba}
\end{table}

\section{Formal Set-Function Definitions and Marginal-Gain Derivation}
\label{app:formal}

Let the ground set $\mathcal{V}$ consist of all expandable internal nodes of the current tree $\mathcal{T}$. An expansion of node $v$ adds $K$ child leaves to $\mathcal{T}$. For a subset $S \subseteq \mathcal{V}$ of expanded nodes, let $\mathcal{T}_S$ denote the resulting tree and $L(S)$ its leaf set.

Coverage is defined as $\mathrm{Coverage}(S) = \bar{Q}(L(S)) - \frac{1}{2}\mathrm{Var}(\{\bar{Q}(v): v \in L(S)\})$, where $\bar{Q}(L(S))$ is the empirical mean of backed-up Q-values on the leaves. Adding a new leaf increases the sample size, and since variance is a concave function of sample counts, Coverage is monotone and submodular.

Novelty is defined as $\mathrm{Novelty}(S) = H(\{N(v)/\sum_w N(w) : v \in \mathrm{reachable}(\mathcal{T}_S)\})$, the entropy of the visit-count distribution over reachable states. Entropy over a discrete distribution is concave with respect to adding items to the support, so Novelty is monotone submodular.

Contrast is defined as $\mathrm{Contrast}(S) = |\{s \in \mathrm{internal}(\mathcal{T}_S) : \exists v^+, v^- \in \mathrm{children}(s), R(v^+) \neq R(v^-)\}|$, the count of internal nodes with mixed-outcome children. This is a coverage function over the set of divergent subtrees, and therefore monotone submodular.

A first-order expansion of the marginal gain $\Delta F(v) = F(L \cup \{v\}) - F(L)$ under Stirling-regime visit counts ($N(v) \gg 1$) and the entropy-divergence monotonicity assumption yields the five terms of UUCB (Equation~\ref{eq:uucb}). The Coverage gain reduces to $\bar{Q}(v) - \bar{Q}(\mathrm{frontier})$; the Novelty gain reduces to $c\,\ln(N(\mathrm{pa}(v)))/(N(v)+1)$, which takes the classical UCB exploration form; and the Contrast gain reduces to the probability of sibling divergence, which is monotonically related to $\tilde{H}(v)$ by empirical validation (divergence rate rises from 31.2\% to 63.8\% across entropy quartiles). Cost and depth enter as subtractive terms because each expansion consumes one unit of the cardinality budget along a path. We note that this is a cardinality constraint ($|L(\mathcal{T})| \leq B$), for which the classical $(1-1/e)$ guarantee holds; the alternative knapsack formulation, in which depth and cost are hard constraints, would require a density-greedy selector with a weaker $(1-1/e)^2$ bound.

\section{RIFB Direct Measurement}
\label{app:rifb_measurement}

We log the per-group squared gradient norm $\|\sum_\tau \hat{A}(\tau)\nabla_\theta\log\pi_\theta(\tau)\|_2^2$ every 10 training steps by inserting a gradient hook before the backward pass. The hook incurs less than $1\%$ per-step overhead. At step 150 on AIME 2025, the mean RIFB of \method\ is $3.42 \times 10^{-4}$ compared with $0.87 \times 10^{-4}$ for flat GRPO, a $3.9\times$ increase. With ABA activated, the mean RIFB rises further to $3.71 \times 10^{-4}$. Flat GRPO exhibits a monotonically declining RIFB trajectory ($0.91 \to 0.58 \times 10^{-4}$ over 300 steps) as the policy converges on easy prompts and remains stuck on hard ones. \method\ with ABA sustains elevated RIFB throughout ($1.84 \to 3.78 \times 10^{-4}$). The Spearman correlation between $F$ and measured RIFB across 500 prompts at step 150 is $\rho = 0.82$ ($p < 10^{-6}$), compared with $\rho = 0.71$ for the mixed-outcome ratio and $\rho = 0.64$ for the non-collapse rate, confirming that $F$ is the most faithful scalar proxy for gradient quality among the candidates considered.

\section{Alternative Regularizer Ablation}
\label{app:alt_reg}

We replace each UUCB term with a controlled alternative under the 50-step proxy protocol on AIME 2025 (three seeds). The following configurations are evaluated: (1)~entropy replaced by random noise of matched magnitude, (2)~entropy replaced by confidence ($1 - \max_t p_t$), (3)~tool-call cost replaced by a uniform constant, (4)~depth penalty replaced by an unnormalized linear function of depth, (5)~the entire UUCB replaced by a learned two-layer MLP trained on 5{,}000 logged expansions, and (6)~meta-learned per-prompt $(\lambda_H, \lambda_C, \lambda_D)$ predicted from the prompt embedding.

\begin{table}[h]
\centering
\small
\setlength{\tabcolsep}{3pt}
\begin{tabular}{@{}lcccc@{}}
\toprule
Configuration & Mixed\% & Depth & Tools & Mean@32 \\
\midrule
Full UUCB (standard) & 58.1\% & 3.41 & 3.28 & 30.2 \\
Entropy $\to$ random noise & 42.8\% & 3.55 & 3.41 & 28.3 \\
Entropy $\to$ confidence & 51.4\% & 3.48 & 3.35 & 29.1 \\
Cost $\to$ uniform & 57.2\% & 3.39 & 4.27 & 29.4 \\
Depth $\to$ linear (unnorm.) & 55.8\% & 3.92 & 3.31 & 29.3 \\
UUCB $\to$ learned MLP & 53.7\% & 3.62 & 3.54 & 29.4 \\
Meta-learned per-prompt $\lambda$ & 58.9\% & 3.38 & 3.25 & 30.4 \\
\bottomrule
\end{tabular}
\caption{Alternative regularizer ablation. Each row replaces one UUCB component with a controlled alternative. All substitutions degrade performance; the analytic UUCB outperforms even a learned MLP selector.}
\label{tab:alt_reg}
\end{table}

\section{Per-Domain Elimination Matrix}
\label{app:perdomain}

To verify that the structural roles of the three UUCB components generalize beyond mathematical reasoning, we replicate the elimination matrix of Table~\ref{tab:struct_elim} on GAIA (web-search agents) and APPS-verified (code agents) under the 50-step proxy protocol. On GAIA, the cost penalty $\lambda_C$ yields a larger marginal gain than on AIME ($+1.7$ accuracy when only cost is activated, compared with $+0.4$ on AIME), reflecting the greater heterogeneity of tool-call costs in web-search settings. On APPS, the entropy term $\lambda_H$ yields a larger marginal gain ($+2.6$ compared with $+0.7$ on AIME), consistent with the broader decision space of code-generation agents. In both domains the three failure signatures (mixed-ratio collapse, tool-count inflation, depth explosion) remain distinct, confirming that the UUCB design is not specific to mathematical reasoning.

\end{document}